\documentclass[letterpaper, 10 pt, conference]{ieeeconf}  

\IEEEoverridecommandlockouts                              

 \usepackage{booktabs}      
\usepackage{caption}       
\usepackage{array}         
\usepackage{siunitx}
\usepackage[table]{xcolor}
\usepackage[printonlyused,nohyperlinks]{acronym}
\usepackage{graphicx}
\usepackage{cite}
\usepackage{amsmath}
\usepackage{amssymb}
\usepackage{bm}

\title{\LARGE \bf
Leveraging Self-Supervised Learning Methods for Remote Screening of Subjects with Paroxysmal Atrial Fibrillation
}

\author{
    Adrian Atienza$^{*1}$, 
    Gouthamaan Manimaran$^{*1}$, 
    Sadasivan Puthusserypady$^{1}$,
    Helena Domínguez$^{2}$,\\
    Peter K. Jacobsen$^{3}$
    and Jakob E. Bardram$^{1}$
\thanks{$^{*}$Contributed equally}%
\thanks{*This work was supported by Innovation Fund Denmark}
\thanks{$^{1}$A. Atienza, G. Manimaran, S. Puthusserypady and J.E. Bardram are with the Department of Health Technology, Technical University of Denmark.}%
\thanks{$^{2}$H. Domínguez is with the Department of Cardiology at the University Hospital–Bispebjerg.}%
\thanks{$^{3}$P.K. Jacobsen is with the Department of Heart Diseases at Rigshospitalet.}%
}

\begin{document}

\maketitle
\thispagestyle{empty}
\pagestyle{empty}

\begin{abstract}

The integration of \ac{ai} into clinical research has great potential to reveal patterns that are difficult for humans to detect,
creating impactful connections between inputs and clinical outcomes.
However, these methods often require large amounts of labeled data, which can be difficult to obtain in healthcare due to strict privacy laws and the need for experts to annotate data.
This requirement creates a bottleneck when investigating unexplored clinical questions.
This study explores the application of \ac{ssl} as a way to obtain preliminary results from clinical studies with limited sized cohorts. 
To assess our approach, we focus on an underexplored clinical task: screening subjects for \acf{paf2} using remote monitoring, single-lead ECG signals captured during normal sinus rhythm. 
We evaluate state-of-the-art SSL methods alongside supervised learning approaches,
where SSL outperforms supervised learning in this task of interest.
More importantly, it prevents misleading conclusions that may arise from poor performance in the latter paradigm when dealing with limited cohort settings.


\vspace{0.25cm}

\indent \textit{Clinical relevance}- 
This study illustrates how self-supervised learning (SSL) provides robust preliminary studies with minimal labeled data. By leveraging SSL, researchers can assess the feasibility of clinical questions before committing to extensive data collection efforts. 
In addition, our findings demonstrate that \ac{paf2} can be effectively detected from normal sinus rhythm recordings captured by wearable devices. This capability paves the way for scalable population screening, potentially transforming early diagnosis and intervention strategies in clinical practice.

\end{abstract}

\section{INTRODUCTION}

The influence of \acf{ai} in the medical field has been growing significantly and 
these algorithms are expected to play a significant role in the future of healthcare.
On one hand, AI offers a powerful complementary diagnostic tool that can streamline clinical workflows and serve as an effective means of validation \cite{adelaja2023operating}. Furthermore, AI has become indispensable in clinical research, as these advanced algorithms are capable of detecting complex data patterns that extend beyond human visual capabilities—ultimately uncovering clinically significant insights that may otherwise remain hidden \cite{pettit2021artificial}.
However, traditional AI methods need a large amount of labeled data to identify these patterns in the data.
This data is difficult and expensive to gather \cite{alabduljabbar2024medical}.
This bottleneck restricts the clinical research to outcomes that have been labeled in massive datasets, leaving a lot of clinical questions unaddressed.

\vspace{0.25cm}
The increasing interaction between humans and new technologies results in a massive amount of unlabeled data. 
This trend has been recognized by \ac{ai} researchers across various domains, including healthcare. 
Consequently, methods have been developed to transform these unlabeled sources into usable information. 
Self-Supervised Learning (SSL) techniques have emerged as a result of this effort.
These framework drives AI models to understand data and compute informative representations without the explicit supervision of labels associated to the data. \cite{gui2024survey}. 
It has been successfully applied to all data domains, i.e., Natural Language Processing \cite{kotei2023systematic}, Computer Vision \cite{shurrab2022self} and Time Series \cite{zhang2024self}.
This paper studies the applicability of this promising framework to clinical research.
We hypothesize that these informative label-agnostic representations can enable preliminary studies with extremely limited labelled cohort. 
These studies can offer definitive insights into the feasibility of answering critical clinical questions, thereby directing resources for labeled data collection towards those that can be resolved.

\vspace{0.25cm}
To assess our hypothesis we have evaluated the Self-Supervised Framework in a relevant unsolved clinical question; \textit{Leveraging remote monitoring Single-Lead ECG signals in sinus rhythm condition to screen subjects with \ac{paf2} condition}.
Existing literature \cite{lancet} proves that baseline 12-Lead ECG accommodates information in order to discretize between the control group and the risk group associated with \ac{paf2} subjects.
This study aims to explore screening patients with \ac{paf2} condition from NSR ECG signals captured by remote monitoring devices.
Using signals captured by a remote monitoring devices instead of an in-clinic 10 seconds 12-Lead ECG signal has the potential to scale this population screening since it is common that people nowadays has access to this kind of devices.

\vspace{0.25cm}
We have carried out an extensive evaluation on the \ac{paf} \cite{datasetaf}, which only accommodates 50 subjects.
This is the only publicly available cohort prepared for addressing this clinical question.
The evaluation is structured to derive two distinct conclusions from the analysis of the results:
(i) The potential of SSL for addressing clinical questions when the labeled cohort is limited compared with the standard supervised framework.
(ii) The feasibility of screening subjects with \ac{paf2} from remote monitoring devices.
Accordingly, this evaluation accommodates both the most common deep learning architectures used in supervised learning and the state-of-the-art SSL learning methods tailored for ECG processing. 
Notably, we found a significant gap between the supervised learning and the SSL framework under this extreme data conditions.
While the results from the SSL framework conclusively demonstrate the feasibility of the task, relying solely on the supervised framework might lead to incorrect conclusions and discard its feasibility.

\vspace{0.25cm}

In summary, the contributions of the paper are:
\begin{itemize}
    \item We discuss one of the primary challenges in addressing clinical questions with AI: the necessity for large cohorts.
    
    \item We have demonstrated the advantages of utilizing the SSL framework to achieve robust preliminary results with a limited amount of data.
    
    \item By employing this paradigm, we have shown the feasibility of identifying subjects with AFib through remote monitoring devices. 
\end{itemize}

\section{Literature Review}
Deep learning architectures have been widely used when managing unstructured data such as language, images, or signals. 
These models mimic the brain's processing of such inputs, converting them into structured features for specific tasks.
Each neural architecture has unique characteristics but generally consists of two main components: the encoder, which converts unstructured data into a structured representation, and the discriminative part, which maps this representation to the desired outcome.
This section explores the differences in model optimization between the traditional Supervised Learning approach and the emerging Self-Supervised Learning paradigm.

\vspace{0.25cm}

\textbf{Supervised Learning}
Supervised Learning has been the framework par excellence when it comes to optimizing deep learning architectures to solve a specific purpose.
During the optimization process, the model is provided with pairs of inputs and labels, which involves a single step. 
The input, the ground truth label, and the predicted label computed by the model are combined in a standard loss function used for optimizing both the encoder and the discriminative part simultaneously.
Given that the target function is standardized, advancements in this field primarily concentrate on the design of the deep learning architecture employed.
This approach has yielded excellent results across various data types. However, its primary limitation is its significant reliance on the volume of data in the dataset.
The dataset must provide enough variability to allow the model to identify generalizable patterns for the intended task.

\vspace{0.25cm}

\textbf{Self-Supervised Learning}
The primary distinction between this framework and Supervised Learning is that the optimization of the encoder and the discriminative part occurs separately.
The encoder is optimized using massive unlabelled datasets. 
A training goal is set to ensure the model comprehends the data and captures its complexity within the computed representations. 
After this process, the discriminative part is optimized for the specific task using the representations computed by the already pre-trained encoder whose parameters remain fixed.
Since the encoder contains most of the optimizable parameters, it is anticipated that a significantly smaller amount of annotated data will be required to successfully complete the task.
Recent research indicates that this two-step optimization process yields superior results across various data modalities.

\section{Methods}
This section provides a more detailed description of the characteristics of the existing state-of-the-art methods that are going to be evaluated for the screening of \ac{paf2} task.
We have split them into two groups, i.e., Contrastive Learning Family and other methods.
\label{sec:methods}

\subsection{Contrastive Learning Family}
This framework was introduced by SimCLR \cite{simclr}. 
In its original formulation, it relies on creating two versions of the same input using data augmentation techniques.
The two pairs of inputs (positive pairs), are processed using the same encoder.
The model is driven to compute similar representations for the two positive pairs, while keeping these representations distant from other pairs in the batch (negative pairs).
This training objective is described in the following loss function:

\begin{equation}\label{eq:cl}
\ell_{i, j}=-\log \frac{\exp \left(\operatorname{sim}\left(\boldsymbol{z}_i, \boldsymbol{z}_j\right) / \tau\right)}{\sum_{k=1}^{2 N} \mathbb{1}_{[k \neq i]} \exp \left(\operatorname{sim}\left(\boldsymbol{z}_i, \boldsymbol{z}_k\right) / \tau\right)},
\end{equation}

Although its original intended use was for image processing, it has been leveraged for processing other kinds of data, including ECG signals. The methods described below have their own uniqueness, but all of them utilizes this approach.

\vspace{0.5cm}

\textbf{Mix-UP \cite{wickstrom2022mixing}:} 
This method introduces a more tailored data augmentation for time series, where two non-overlapping strips are merged with weights sampled from a beta distribution. 
The loss function described in Eq.~\ref{eq:cl} is computed between each of the two strips and the corresponding version product of their combination.
\vspace{0.25cm}

\textbf{CLOCS \cite{kiyasseh2021clocs}:} This method proposes to lead the model to encode subject-unique patterns invariant to short-time data shifts. The model is driven to compute similar representations by computing Eq.~\ref{eq:cl} over two consecutive ECG strips sampled from the same recording. 

\vspace{0.25cm}

\textbf{PCLR \cite{pclr}:} This method pushes the model to encode the subject-unique characteristics that are invariant to long-term data shifts.
For this purpose, Eq~\ref{eq:cl} is computed between ECG strips sampled from different recordings of the same patient with a large temporal gap between them.

\vspace{0.5cm}

\subsection{Other Methods}
Although Contrastive Learning is the most commonly used SSL framework in the realm of ECG processing, recently other methods have been presenting alternatives to the use of this framework with a high level of success. These methods are described bellow:

\vspace{0.25cm}

\textbf{DEAPS \cite{deaps}:}
This method adapts a non-Contrastive Learning technique. 
This means that while the model is still encouraged to compute similar representations between positive pairs, this framework penalises that these representations are distant with respect to the other inputs in the batch. 
This purpose is to incorporate another component to the cost function that drives the model to encode patterns that vary across the recording by encouraging differences between triplets of strips sampled from the same recording.

\vspace{0.25cm}

\textbf{MTAE: \cite{mtae}} This method extends successful masked data modelling methods such as MAE to be adapted for ECG processing. For this purpose, the input signal is masked with a fixed masking ratio. The encoder is driven to compute meaningful patch representations in such a way that they can be used by a decoder to reconstruct the missing portions of the input.

\vspace{0.25cm}
\textbf{NERULA \cite{nerula}:} This method combines generative and discriminative pathways for self-supervised learning task. The discriminative pathway involved non-contrastive learning of two versions of a single masked signal where the second mask is the inverse of the first mask. The generative pathway tries to reconstruct the first masked signal into the ground-truth ECG signal.

\vspace{0.5cm}

\section{Evaluation} \label{sec:eval}
This section outlines the specific datasets utilized for pre-training the SSL methods and the dataset used for evaluating the task of interest.
Additionally, it provides details on the experimental design and the results obtained.
Finally, we discuss the implications of the evaluation results.

\begin{table*}[ht]
    \centering
    \caption{Performance Metrics for Self-Supervised Learning Models}
    \label{tab:modelssl}
    \renewcommand{\arraystretch}{1.2} 
    \resizebox{0.95\textwidth}{!}{ 
    \begin{tabular}{lccc|ccc|ccc}
        \toprule
        & \multicolumn{3}{c}{\textbf{SVC}} & \multicolumn{3}{c}{\textbf{Random Forest}} & \multicolumn{3}{c}{\textbf{Logistic Regression}} \\
        \cmidrule(lr){2-4} \cmidrule(lr){5-7} \cmidrule(lr){8-10}
        \textbf{Model} & \textbf{Accuracy} & \textbf{F1 Score} & \textbf{AUC} & \textbf{Accuracy} & \textbf{F1 Score} & \textbf{AUC} & \textbf{Accuracy} & \textbf{F1 Score} & \textbf{AUC} \\
        \midrule
        \textbf{CLOCS} \cite{kiyasseh2021clocs} & 0.667 ± 0.115 & 0.675 ± 0.111 & 0.672 ± 0.116 & 0.762 ± 0.056 & 0.758 ± 0.056 & 0.745 ± 0.103 & 0.665 ± 0.078 & 0.673 ± 0.075 & 0.727 ± 0.050 \\
        \textbf{DEAPS} \cite{deaps} & 0.701 ± 0.045 & 0.703 ± 0.049 & 0.724 ± 0.073 &  0.747 ± 0.054 & 0.737 ± 0.058 & 0.742 ± 0.068 & 0.616 ± 0.077 & 0.625 ± 0.073 & 0.656 ± 0.083 \\
        \textbf{MTAE} \cite{mtae} & 0.710 ± 0.083 & 0.713 ± 0.080 & 0.767 ± 0.080 & 0.731 ± 0.057 & 0.728 ± 0.061 & 0.791 ± 0.076 & 0.697 ± 0.078 & 0.700 ± 0.072 & 0.734 ± 0.056 \\
        \textbf{NERULA} \cite{nerula} & 0.668 ± 0.046 & 0.671 ± 0.041 & 0.749 ± 0.109 & 0.773 ± 0.088 & \textbf{0.771 ± 0.086} & 0.824 ± 0.107 & 0.700 ± 0.110 & 0.700 ± 0.104 & \textbf{0.792 ± 0.131} \\
        \textbf{Mix\_Up} \cite{wickstrom2022mixing} & 0.750 ± 0.038 & 0.751 ± 0.040 & 0.786 ± 0.093 & 0.765 ± 0.063 & 0.758 ± 0.063 & 0.794 ± 0.107 & 0.643 ± 0.088 & 0.650 ± 0.084 & 0.691 ± 0.102 \\
        \textbf{PCLR} \cite{pclr} & \textbf{0.782 ± 0.038} & \textbf{0.774 ± 0.043} & \textbf{0.808 ± 0.085} & \textbf{0.774 ± 0.040} & 0.751 ± 0.050 & \textbf{0.825 ± 0.095} & \textbf{0.714 ± 0.049} & \textbf{0.711 ± 0.054} & 0.732 ± 0.091 \\
        \bottomrule
    \end{tabular}
    }
\end{table*}

\subsection{Description of the Datasets}
\vspace{0.25cm}
\paragraph{Pre-training Dataset} 
The various SSL methods evaluated necessitate different data specifications. For instance, PCLR requires time strips with extended time gaps, whereas DEAPS needs recordings longer than 2 minutes.
To achieve the primary goal of evaluating SSL methods independently of the database, all methods have been pre-trained using the Sleep Heart Health Study (SHHS) \cite{shhs}.
This cohort has been identified as the only large publicly available dataset that meets the conditions for all methods used in the evaluation.
Being a multi-center cohort study which involves 6,441 participants, it accomodates ECG recordings alongside other physiological signals collected during overnight sleep across two visits, spaced three years apart.

\vspace{0.25cm}

\paragraph{Evaluation Dataset} For our evaluation experiment which focuses on \acf{afib} screening, we utilized the \acf{paf} dataset \cite{datasetaf}, which contains data from 50 patients. Of these, 25 participants exhibit \ac{paf2} and have two 30-minute ECG recordings each. The first recording is taken during a period distant from any \ac{afib} episode, while the second is captured immediately before an episode. In our experiments, we consider only the first recording from each patient, as they represent pure \ac{nsr} without pre-\ac{afib} markers, thereby better simulating a screening scenario.

\vspace{0.25cm}

\paragraph{Data Processing}
All signals were uniformly sampled at 100\,Hz, with each 10-second segment treated as an individual input instance. 
Both pre-training and evaluation signals were preprocessed using a second-order Butterworth bandpass filter, restricting the frequency range to 0.5--40\,Hz.

\subsection{Experimental Design}
Given the limited number of participants in the evaluation dataset, we employed stratified 5-fold cross-validation with patient-wise splits to prevent data leakage.
We ensure balanced class representation across folds. 
Our experiments were implemented in \texttt{PyTorch} and executed on a single NVIDIA Tesla V100 GPU. 
For all the experiments, the standard accuracy, F1 score, and area under the ROC curve (AUC) metrics are computed for a comparison of the results. 
\vspace{0.5cm}

\subsection{Baselines}

\vspace{0.25cm
}
\textbf{Supervised-Learning Architectures:}
We compare the self-supervised methods with some well-known supervised models widely used for ECG processing, i.e., Res-Net50 \cite{resnet}, Inception \cite{inception}, Vision Transformer \cite{vit}, EfficientNet-v1 \cite{efficientnet}, EfficientNet-v2 \cite{efficientnetv2}, Local-Lead Attention \cite{locallead} and Swin Transformers \cite{swin}. 
All these methods were trained in a supervised fashion on the evaluation dataset, 
utilizing the binary cross-entropy loss and Adam optimizer with an initial learning rate of \(1 \times 10^{-3}\). 
To further enhance training efficiency, we applied a cosine annealing learning rate scheduler.
The training process concludes when the loss function value ceases to decrease in the validation set, in line with standard supervised learning practices.
  
\vspace{0.25cm}
\textbf{Self-Supervised Learning Methods:}
Initially, a Logistic Regression \cite{logisticreg} model was fitted on top of the representations computed by the pre-trained encoder. 
This replicates the discriminative part of deep learning architecture trained using Supervised Learning. 
It is worth mentioning that the pre-trained encoder does not necessitate the optimization of this machine learning model to be differentiable for gradient propagation to the encoder.
It additionally allows us to fit other machine learning models like Support Vector Classifier (SVC)\cite{svc} or Random Forest \cite{randomforest}. Being more powerful, they are anticipated to yield better results.

\vspace{0.5cm}

\subsection{Discussion of the Results}
As shown in Table \ref{tab:modelssl}, SSL methods achieve a degree of performance that allows us to conclude that it is viable to screen subjects with chornical \ac{paf2} using remote sensors. 
In addition, SSL methods consistently outperformed traditional supervised learning approaches. These latter results are displayed in Table \ref{tab:modelsupervised}.
Results show an improvement of  15.4\% in AUC, alongside increases of 11.6\% in F1 score and 12.5\% in accuracy when compared with the best Supervised Learning results. 
We emphasize that solely relying on Supervised Learning results might lead to the incorrect conclusion that this clinical question is not feasible.
These findings provide robust evidence in favor of the hypotheses posited by this study:
The SSL framework allows for meaningful preliminary studies even with a very limited labeled cohort.
Additionally, it helps avoid incorrect conclusions that might arise from interpreting Supervised Learning results when they are consequence of the lack of a massive labeled cohort. 

\begin{table}[ht]
    \centering
    \caption{Performance Metrics of Supervised Models}
    \label{tab:modelsupervised}
    \resizebox{0.5\textwidth}{!}{ 
    \begin{tabular}{lccc}
        \toprule
        & \multicolumn{3}{c}{\textbf{Performance Metrics}} \\
        \cmidrule(lr){2-4}
        \textbf{Model} & \textbf{Accuracy} & \textbf{F1 Score} & \textbf{AUC} \\
        \midrule
        \textbf{ResNet50} \cite{resnet} & 0.492 ± 0.209 & 0.475 ± 0.221 & 0.510 ± 0.211 \\
        \textbf{Inception} \cite{inception} & 0.552 ± 0.154 & 0.549 ± 0.149 & 0.645 ± 0.174 \\
        \textbf{Vision Transformer} \cite{vit} & 0.562 ± 0.106 & 0.563 ± 0.107 & 0.564 ± 0.100 \\
        \textbf{ResNeXt} \cite{resnext} & 0.566 ± 0.166 & 0.569 ± 0.160 & 0.624 ± 0.208 \\
        \textbf{EfficientNet-v1} \cite{efficientnet} & 0.579 ± 0.129 & 0.582 ± 0.124 & 0.631 ± 0.183 \\
        \textbf{Efficientnet-v2} \cite{efficientnetv2} & 0.586 ± 0.156 & 0.588 ± 0.147 & 0.632 ± 0.205 \\
        \textbf{Local-Lead} \cite{locallead} & 0.629 ± 0.062 & 0.633 ± 0.055 & 0.698 ± 0.092 \\
        \textbf{Swin Transformer} \cite{swin} & \textbf{0.656 ± 0.088} & \textbf{0.657 ± 0.082} & \textbf{0.670 ± 0.101} \\
        \bottomrule
    \end{tabular}
    }
\end{table}

\section{Leveraging Long-Term Remote Monitoring}

\begin{figure}[!ht]
\centering

\fbox{\includegraphics[width=0.95\columnwidth]{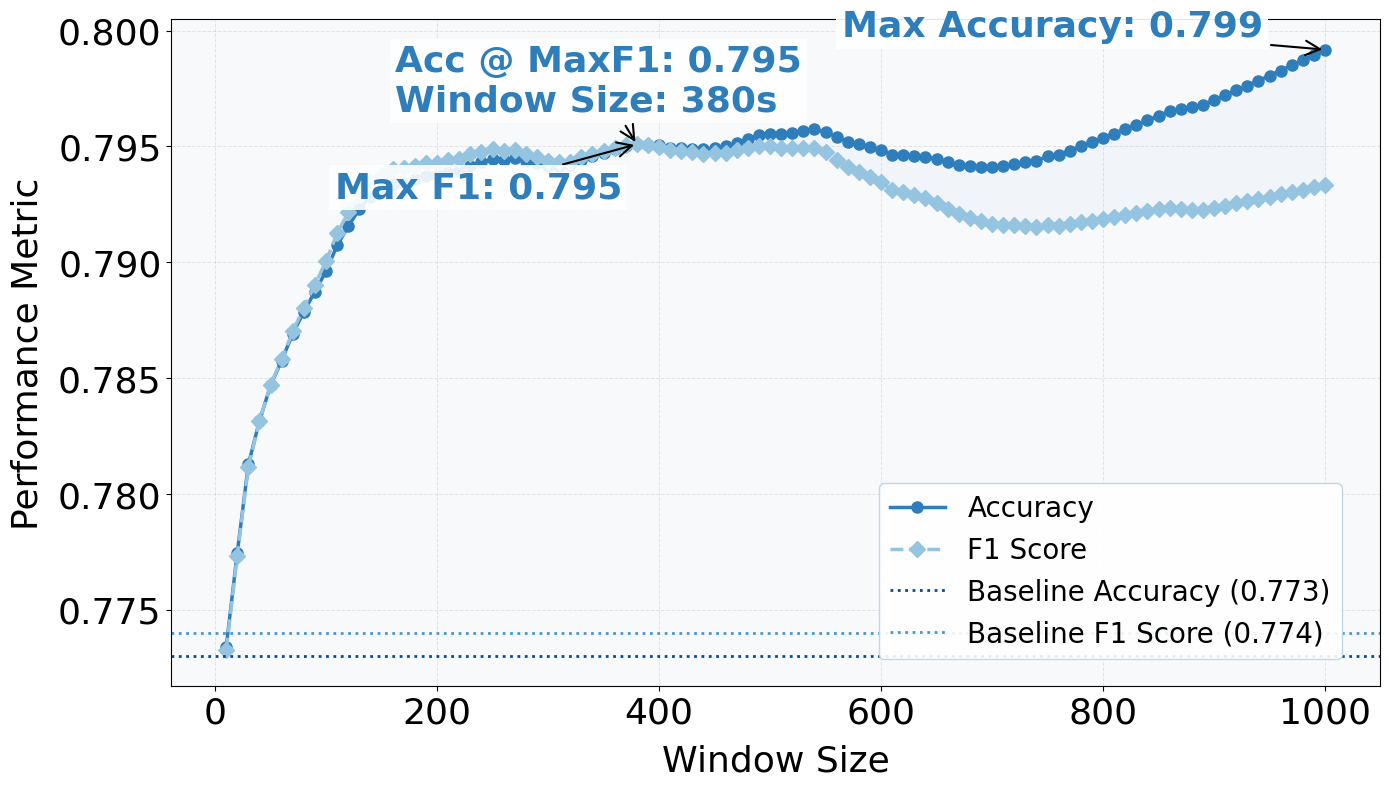}}%
\caption{Accuracy and F1 Score vs Window Size}
\label{fig:window}
\end{figure}

The results presented in section \ref{sec:eval} consider 10-second ECG signals independently, aligned with standard practices. 
However, due to the nature of the remote sensors used to capture cardiac signals, it is highly likely that recordings longer than 10 seconds are available.
This section demonstrates how leveraging this inherent feature of remote monitoring can enhance the results by using time strips longer than 10 seconds for inference. Figure~\ref{fig:window} shows the predicted label with the highest confidence for the final predictions among the \ac{ecg} strips within the given window size.
We observe that while both accuracy and F1 scores increase with longer time lengths, the F1 score drops after a certain point, even though accuracy continues to rise. Therefore, we select the window size $W$ that corresponds to the maximum F1 score, which is $W=380$ seconds. 


\section{Conclusions}
In this study, we have tackled one of the major challenges in using AI for clinical research: the need for large cohorts.
In this context, we have proposed the use of the SSL paradigm to obtain informative preliminary results with very limited cohorts. This approach allows us to determine which clinical questions can be addressed with AI techniques before committing resources to cohort collection.
To illustrate the benefits of the SSL framework, we focused on a critical clinical question: screening patients with \ac{paf2} using single-lead ECG recordings in NSR conditions. 
Solving this challenge has the potential of scaling the screening of the population suffering from this particular disease since these signals are obtained from remote monitoring devices.
We have evaluated the viability of this clinical question on the \ac{paf} cohort, which is extremely limited with just 50 subjects.
From the extensive evaluation conducted, we would like to emphasize two key points: 
(i) The findings confirm that AI techniques, particularly those utilizing pre-trained models with SSL, are capable of addressing this clinical question effectively. 
(ii) Without the inclusion of the SSL paradigm, the feasibility of this approach would be doubtful, as the results from the traditional Supervised Learning paradigm are inconclusive.

\section*{Acknowledgments} 
This research has been funded by the Innovation Fund Denmark as part of the CATCH project (Project No.~\#1061-00046B) and the Copenhagen Center for Health Technology (CACHET).

\newpage 
\section{Acronyms}
\begin{acronym}[E-JEPA]
    \acro  {cvd}   [CVD]   {Cardiovascular Diseases}
    \acro  {afib}  [AFib]   {Atrial Fibrillation}
    \acro  {nsr}  [NSR]   {Normal Sinus Rhythm}
    \acro  {ecg}   [ECG]  {Electrocardiogram}
    \acro   {ai}  [AI] {Artificial Intelligence}
    \acro   {ml}  [ML] {Machine Learning}
    \acro{ssl} [SSL] {Self-Supervised Learning}
    \acro{sgd} [SGD] {Stochastic Gradient Descent}
    \acro{ema} [EMA] {Exponential Moving Average}
    \acro{clocs}[CLOCS]{Contrastive Learning of Cardiac Signals}
    \acro{pclr}[PCLR]{Patient Contrastive Learning}
    \acro{sbncl}[SBnCL]{Subject-Based non Contrastive Learning}
    \acro{sota}[SOTA]{State-of-the-Art}
    
    \acro{byol}  [BYOL] {Boostrap Your Own Latent}
    \acro{vit}  [ViT] {Vision Transformer}
    \acro{ejepa}  [E-JEPA] {ECG-Based Joint-Embedding Predictive Architecture}
    \acro{paf2} [P-AF] {Paroxysmal Atrial Fibrillation} 
    \acro{paf} [PAF] {Paroxysmal Atrial Fibrillation Prediction Challenge Database} 
        
\end{acronym}

\bibliography{root} 
\bibliographystyle{ieeetr}

\end{document}